\documentclass[runningheads]{llncs}
\usepackage{amsfonts}
\usepackage{graphicx}
\usepackage{amsmath}
\usepackage[utf8]{inputenc}

\title{Do Artificial Intelligence Systems Understand?}

\author{Eduardo C. Garrido-Merchán, Carlos Blanco}
\date{July 2022}

\institute{Universidad Pontificia de Comillas, Madrid, Spain
\email{ecgarrido@icade.comillas.edu} \and
Universidad Pontificia de Comillas, Madrid, Spain
\email{cbperez@comillas.edu}}

\begin{document}

\maketitle

\begin{abstract}
    Are intelligent machines really intelligent? Is the underlying philosophical concept of intelligence satisfactory for describing how the present systems work? Is understanding a necessary and sufficient condition for intelligence? If a machine could understand, should we attribute subjectivity to it? This paper addresses the problem of deciding whether the so-called "intelligent machines" are capable of understanding, instead of merely processing signs. It deals with the relationship between syntaxis and semantics. The main thesis concerns the inevitability of semantics for any discussion about the possibility of building conscious machines, condensed into the following two tenets: "If a machine is capable of understanding (in the strong sense), then it must be capable of combining rules and intuitions"; “If semantics cannot be reduced to syntaxis, then a machine cannot understand." Our conclusion states that it is not necessary to attribute understanding to a machine in order to explain its exhibited “intelligent” behavior; a merely syntactic and mechanistic approach to intelligence as a task-solving tool suffices to justify the range of operations that it can display in the current state of technological development. 
\end{abstract}


\section{Introduction}
The intelligence of an agent, which can be a person or a bot, is a subjective property whose definition is unclear and heavily depends on the community that has studied it \cite{legg2007collection}. The main problems of studying this property, from a scientific point of view, are two. First, to provide a culturally and anthropocentrically unbiased measure of it \cite{hilliard1979standardization}; second, that applying this measure can only quantify the behavior shown by the agent but not the potential intelligence that the agent internally has \cite{das2019rules}. Even if we assume that the metric used to measure intelligence in general (which, for example in the case of the intelligence quotient, is very controversial), when one tries to design a test battery to measure the intelligence of an individual the result can be drastically biased as a consequence of the personality of the individual. For instance, consider an individual with a double exceptionality condition, both having an extreme non-verbal autistic syndrome disorder (ASD) and high capacities. This individual would have a high IQ level according to several tests but, probably, his condition will not allow him to complete the test, even to understand it as the rest of the people. However, its analytical skills may excel those of other individuals. Likewise, one may find more examples of this problem regarding memory and the animal kingdom. Consequently, some psychologists argue that the only thing that we can measure is external behavior. However, the computer science community tries to define intelligence within the range of an objective analytical expression.  

Given these limitations, this paper is focused on what can be defined as computational intelligence, not emotional intelligence, human-like intelligence, or general intelligence, which are broader terms, of undoubted psychological relevance. Concretely, the computational intelligence of a system would be informally defined as the ability of an agent to learn how to efficiently and accurately solve a specific task, such as planning, regression, or classification, by having access to data or experimental observations. Interestingly enough, we will show how an intelligent agent, according to this definition of intelligence, does not need to understand a problem in order to solve it. Thus, we will explore a view of machine intelligence devoid of psychological factors, in whose framework we will try to explain the behavior of these systems without invoking an intrinsic faculty of understanding, and therefore a set of mental states. The key concept is that a computational problem can be solved by computing a sequence of steps, an algorithm, that can be inferred from data without needing to understand what the data or the task means.   

This paper is divided into two parts. The first section examines the present state of Artificial Intelligence, paying special attention to its historical development and the possibilities displayed by the most sophisticated tools so far designed, like Deep Learning. Through a combination of technical and non-technical language, it explains the fundamentals of Artificial Intelligence and its inferential machinery, including the most recent innovations in the field. The second section is philosophical. It addresses the conceptual problems posed by the properties of current intelligent machines, and it relies on the basic distinction between syntaxis (as sequential, sign processing following a set of rules) and semantics (as sign understanding, a process that demands “intuition” rather than rules).   

\section{Current Artificial intelligence approaches and models}
Artificial intelligence was a term originally coined at the 1956 Dartmouth conference to describe programs whose behavior mimics that of human beings, considered intelligent. Since those years, several approaches have been used to generate artificial intelligence in a system. Without loss of generality, we split those approaches into four categories: expert systems using a logic inference engine, machine learning and deep learning models and their variations, neurosymbolic artificial intelligence, and, lastly, probabilistic graphical models, and causality.   

\subsection{Expert systems, Good-old artificial intelligence}
Expert systems, nowadays known as Good-Old Fashioned Artificial Intelligence (GOFAI), are knowledge bases filled with logic predicates and atoms. A knowledge base is a series of instructions that a knowledge engineer introduces to the system (whether an atom or a rule), with an antecedent and a consequent. More technically, those are Horn clauses. If the consequent is true, then the antecedent is also true. A logic programming language such as Prolog \cite{clocksin2003programming} incorporates an inference engine that can read all the content of the knowledge base to solve queries put by the practitioner. Finally, the program can be used by a non-expert user to solve queries introduced with a user interface.   

\begin{figure}[h]
    \centering
    \includegraphics[width = 1.0\textwidth]{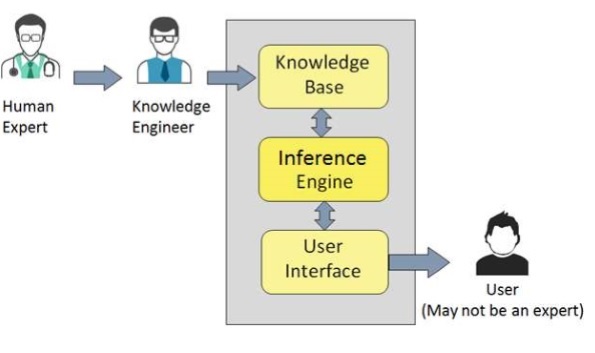}
\caption{Visualization of the modules and information flux of an expert system.}
    \label{fig:gofai}
\end{figure}

More technically, the inference engine uses the Selective Linear Definite clause resolution algorithm (SLD algorithm) to solve the query with the data previously introduced in the knowledge base. The SLD algorithm performs backward reasoning from the query selected by the user creating a search tree of alternative computations to explore the knowledge base, where the query is associated with the root of the tree. The behavior displayed by this kind of system is performing a breadth-first search or, in some cases, more complex search, on a database. In other words, expert systems are database systems with a different search procedure and knowledge representation that the one used in relational (SQL) or document databases like MongoDB. 

A trivial example of an expert system is the following one: We can have an atom representing that a dog is an animal, like \texttt{animal(dog)}. Then, we can have a rule saying that every animal is a living being like 
\begin{align}
\texttt{animal(X):-living\_being(X)} \,.
\end{align} 
Hence, if the user introduces a query looking for a living being like 
\begin{align}
\texttt{:-living\_being(Y)}\,.
\end{align}
Then, the system will retrieve \texttt{Y=dog}. Evidently, this approach presents multiple limitations. First, it does not account for uncertainty. In particular, the majority of relations between elements present uncertainty. For example, if I speak Spanish I may live in Spain, but I may also live in the United States. Fuzzy logic \cite{zadeh1988fuzzy} targets this issue introducing fuzzy causal relations between predicates. However, fuzzy expert systems do not adapt to changes in the environment. If Spanish is no longer spoken in the United States in the next century, the system will not be able to readapt to this reality once it is programmed in a certain way. Most critically, a fuzzy expert system is essentially only executing if-else statements and generating random numbers to verify whether a query is true or not. 

In essence, there is no understanding here, no “internal assimilation” referred to a subject. Knowledge is hard-coded by the user; it is not adaptative to a change in the context of the problem. Expert systems are essentially databases whose queries are solved using the SLD algorithm. However, the behavior shown by expert system applications can be considered quite complex as more and more logic predicates and atoms are introduced into the database. Yet, without the presence of the knowledge engineer to maintain the knowledge base, the expert system is completely unable to interact with the environment, nor can it adapt to changes in it. Consequently, it does not display intelligent behaviour, as it is unable to solve any task that has not been hard-coded in the system through a clearly defined set of instructions.   

\subsection{Machine learning and deep learning models}
As we have seen, expert system models do not adapt to changes in reality. Thus, it is unfeasible to work with them to solve certain tasks such as natural language processing of social networks. In this particular example, new significants S are born every year to reference the same meaning m. For example, Twitter users write “Whiskey” in more than 100 different forms (Wisky, guisky, guiski, güiski...). Eventually, some ways of saying “Whiskey” are not used anymore as time flows. Most critically, if we would like to, for example, detect irony in such texts, the task would become even more difficult, as the variables involved can be all the words in the dictionary plus all the ways of writing those words not included in the dictionary but used by the speakers, plus all the possible combinations of an order of these words, being syntactically correct or not, as Twitter users do not necessarily write syntactically correct texts.  

The previous example, where the task to be solved lies in classifying the value of a categorical variable or performing regression of a continuous variable concerning other variables, presents huge problems to GOFAI, because the number of independent variables is very high (the dimensionality of the problem is very high, where each variable is considered a dimension). In consequence, the artificial intelligence community switched to using statistical learning techniques such as linear or logistic regression. In particular, these techniques associate a real-valued parameter $b$ to each independent variable $x$ to predict and explain the dependent variable $y$. In the case of linear regression, $y = \sum \beta \textbf{x} + \textbf{u}$, these values can be computed with an analytical-closed expression that gives the optimal values $\beta$ that minimize the errors, or residuals, committed by the predictor with respect to a particular sample of data.  

An example is to predict the salary of an employee ($y$) knowing their years of experience at a company ($x_1$), education level ($x_2$), working hours ($x_3$), or company sector ($x_4$). The relation of the salary concerning the other variables is encoded in the values of the ($\beta_1,\beta_2,\beta_3,\beta_4,$ and $\beta_5$) parameters and in the expression $y = \sum \beta \textbf{x} + \textbf{u}$ that creates a linear dependence between the $y$ and each $x_i$. In particular, these values are fixed according to the data $\mathcal{D}$ retrieved of all the employees of the company $\mathcal{D}=\{(\mathbf{X},\mathbf{y})\}$. As a consequence, if employees change or the company change and salaries too, we would only need to retrain the logistic regression model and it would find the new values of the $\beta$ parameters. In particular, this would be easy to do periodically with software engineering.  

It is interesting to note that machine learning models can be interpreted as probabilistic models. For example, the previous example can be interpreted as the following multivariate normal distribution $\mathcal{N}$ with the following expression $\textbf{y} = \mathcal{N}(\beta \mathbf{X}, \sigma I)$ where the mean vector is $\beta \mathbf{X}$ and the covariance matrix is $\sigma I$. As we will further see, neural networks can also be interpreted in this way, where for example in the case of classification, the last layer may be defined as a Bernoulli distribution whose probability depends on the matrix products of the rest of the layers. If we interpret models in this way, we can also generalize them in a Bayesian way, where each of the parameters of the regression $\beta$, or any other set of parameters $\theta$, can be defined with a prior Gaussian distribution $N(\theta, \sigma)$. The posterior distribution $p(\theta \mid \mathbf{X})$ on the parameters $\theta$ would be defined by this Gaussian prior and a likelihood function of the dataset $D=\{(\mathbf{X},\mathbf{y})\}$ using Bayes theorem. Finally, to make more general predictions than classical machine learning, the predictive distribution on a new input value $\mathbf{x}^\star$ would be given by 
\begin{align}
p(\mathbf{y}^\star\mid \mathbf{X}, \theta, \mathbf{x}^\star) = \int p(\mathbf{y}^\star\mid  \theta, \mathbf{x}^\star) p(\theta\mid \mathbf{X}) d\theta \,,
\end{align}
that is, a weighted sum of the predictions $p(\mathbf{y}^\star\mid  \theta, \mathbf{x}^\star)$ made by a model with parameters $\theta$ where the probability of that model based on data would be $p(\theta\mid \mathbf{X})$. Hence, machine learning can be embedded in the Bayesian framework.   

It is important to justify this vision on machine learning models because, if we can interpret them on a Bayesian framework, representing complex probabilistic graphical models, then, we can justify how GOFAI systems are particular cases of these models. In other words, machine learning generalizes GOFAI systems. Thus, and interestingly from a philosophical point of view, the properties of GOFAI systems would be a subset of those of machine learning models. In a first-order logic system, we can have properties such as if an event A happens, then B happens: $A \to B$. We can represent this relation of logical consequence by using a probabilistic graphical model concerning two random variables $A$ and $B$ causally chained. More concretely, using one conditional probability distribution $p(B\mid A)$, for example, a univariate Gaussian distribution $p(B\mid A, \sigma)$. If we set $\sigma=0$ and $A$ models a dichotomous variable with a value true represented as a 1, $p(B\mid A=1, 0)$, then $B$ would be 1. Similarly, $p(B\mid A=0, 0)=0$. Hence, we can embed first-order logic with broader probabilistic graphical models. More interestingly, these models are much more flexible, as we the entire set of known parametric distributions with parameters belonging to the set of real numbers and even non-parametric distributions that can be estimated with algorithms such as kernel density estimators. Consequently, GOFAI systems are a subset of machine learning models, and hence, their properties are a subset of the ones of machine learning models. Moreover, as we have seen, both systems can be reduced to a sequence of computational steps that are eventually executed as binary code in a set of processors. As a consequence, both frameworks are instances of a universal Turing machine. Therefore, its properties are also a subset of those of a universal Turing machine, and can therefore be comprehended as instantiations of algorithmic processing, without the need of invoking some kind of semantic dimension.

Coming back to a description of machine learning models, the majority of them are more complex than linear regression and lack an analytical close expression to obtain the best values of their parameters. On the other hand, those models have a higher capacity than linear regression, assuming fewer hypotheses about the data, and being able to represent more complex functions. For example, linear regression is only able to represent linear relationships between the independent variables and the dependent variable like lines, planes, or hyperplanes restricted by the parametrical function $y = \sum \beta \mathbf{x} + \mathbf{u}$. However, neural networks or Gaussian processes can represent any function $f$ such that $y=f(\mathbf{X}; \theta)$, where $\theta$ is the set of parameters of the neural network or the Gaussian process that are fit to minimize a loss function $\mathcal{L}$ of the predictions of the model $y^\star$ and the real data $y$. Nevertheless, these models are more difficult to fit and offer higher risk of committing dramatic failures on new data because of the overfitting problem.

For example, suppose that we have to predict the amount of information shared in a telecommunication system, $y$, whose number of clients is the $30\%$ market share of a country to build new infrastructure. That could be millions of customers with different services like the Internet, mobile phones whose characteristics, $\mathbf{X}$, like age ($x_1$) or whether they speak at day or night ($x_2$) or the photos that they share ($x_3$) are different non-structured data. Most importantly, a useful pattern may be a complex combination of the values of those variables and those variables can be counted by millions. For instance, a middle-aged person that is present on, at least, $4$ social networks, travels using public transport, and has a high-quality smartphone is predicted to consume more than the average. Those patterns cannot be found automatically by regressions as they assume linearity, $y = \sum \beta \mathbf{x} + \mathbf{u}$, but they can be found using Gaussian processes or neural networks as their models can fit any function such that $y=f(\mathbf{X})$.

When the parameters of statistical models $\theta$ are learned via an iterative optimization algorithm, there is no analytical expression that provides an optimal solution for them. Based on correcting the values of the parameters $\theta$ as a result of a loss function of the error committed by the predictions of the model $\mathcal{L}(\mathbf{y}, \mathbf{y}^\star\mid \theta)$, the computer science community called the family of those techniques “machine learning.” 

In essence, we define the learning process of a machine learning model as follows: an algorithm A that learns from experience E concerning some class of tasks T and performance measure P, if its performance at tasks in T, as measured by P, improves with experience E \cite{lecun2015deep}. Given this abstract description, there are a plethora of machine learning algorithms that use this logic to solve different tasks such as classification or regression. For example, the k-Nearest neighbors algorithm classifies a point according to the values of the data that is more similar to it. The decision tree algorithm chooses the partitions that minimize the entropy of the dataset iteratively to build a set of rules to classify the label of an instance. The support vector machine computes the hyperplane that minimizes the generalization error of the predictions using kernel functions to make the data linearly separable. Recall that all those algorithms fit their parameters according to a given sample of data and have hyper-parameters whose values are chosen by users and generalize their behavior. For example, the similarity function of the k-nearest neighbors algorithm must be set by the user.

A particular machine learning algorithm whose parameter values $\theta$ are also set according to a training iterative optimization algorithm are neural networks. Neural networks were created as a reductionist analogy to the neural networks of biological beings. In particular, the basic architecture of a neural network consists of several neurons that are organized in layers where the links of every neuron with all the rest of the neurons are weighted by the value of a parameter. In essence, every neuron of the neural network is performing a generalization of linear regression $y = \alpha(\sum \beta \mathbf{x} + \mathbf{u})$, where $\alpha$ is a non-linear function, and the outputs of every neuron are transmitted to all the other neurons of the next layer of neurons of the network. When a new instance $\mathbf{x}^\star$ is presented, its values are transmitted into the neural network via a feed-forward algorithm and the network predicts the associated value of the instance $y^\star$. Then, we can compute the error performed by the net $\mathcal{L}(\mathbf{y}, \mathbf{y}^\star\mid \theta)$, and using a backpropagation algorithm based on the classic calculus chain rule of derivatives the weights of the network are reconfigured to adapt themselves to the new instance and minimize the generalization error, or loss function $\mathcal{L}(\mathbf{y}, \mathbf{y}^\star\mid \theta)$, performed by the neural network. 

Several hyper-parameters appear in neural networks such as the number of layers, number of neurons, learning rate, activation functions, optimization algorithms, regularization, and more. Specifically, when the number of hidden layers, those in the middle of the input and output layer, is higher than two, the models are considered to belong to the deep learning class. In particular, these models are universal approximators of functions, that is, they can fit any function $f$, such that $y=f(\mathbf{X}\mid \theta)$, given a representative dataset and architecture of the neural network established by its parameters and hyper-parameters represented by the set of values $\theta$. However, as the model is more complex, it is more difficult to train, as it can suffer from overfitting.    

So far, we have focused our presentation of artificial intelligence models on predictions, showing that any function can be predicted given enough data of its underlying probability distribution $p(\mathbf{X}, \mathbf{y})$, but deep neural networks are also able to generate data y from existing data x. For example, the DALLE-2 model \cite{ramesh2021zero} can generate entirely new photos from text without human intervention. Indeed, it can generate “an astronaut riding a horse in a photorealistic style”. 

\begin{figure}[h]
    \centering
    \includegraphics[width = 0.8\textwidth]{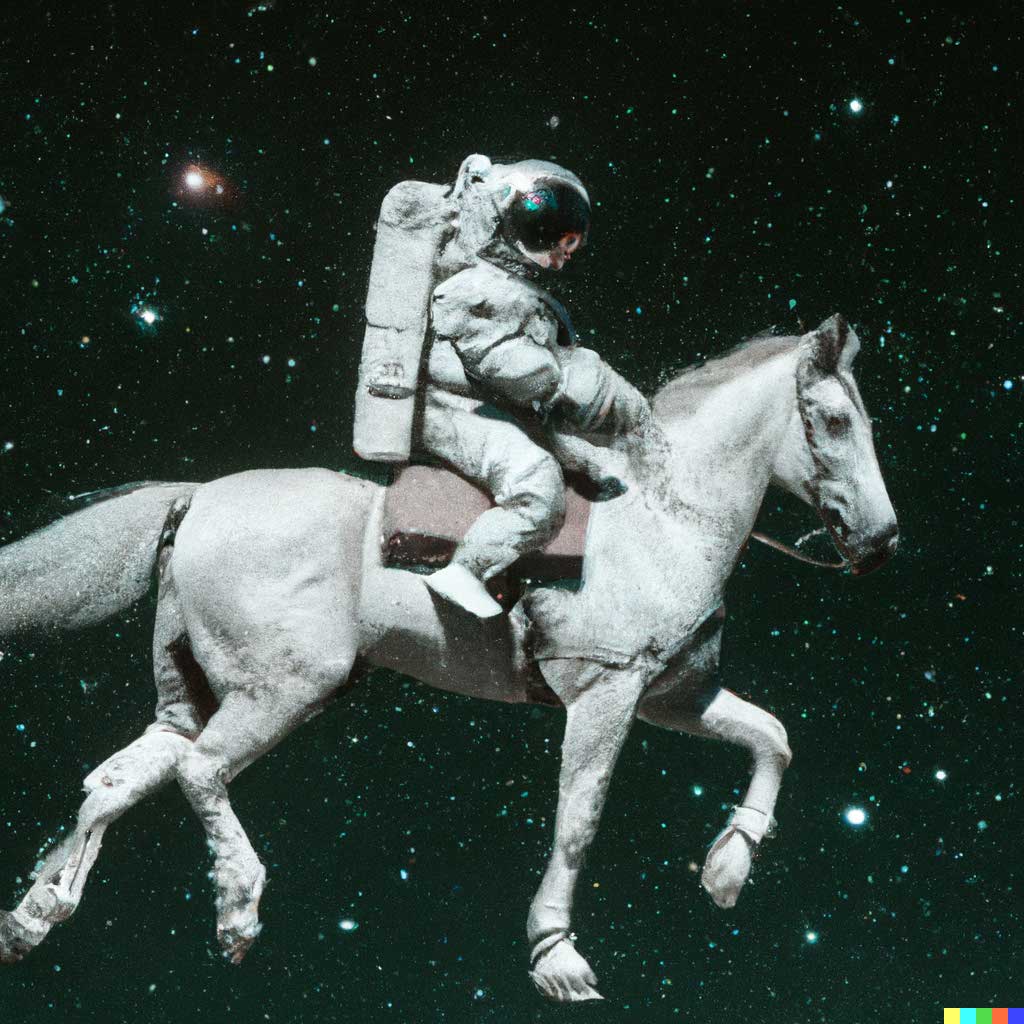}
\caption{Generated photo by DALLE2 \cite{ramesh2021zero}, without human intervention, when we input as a query “an astronaut riding a horse in a photorealistic style.” }
    \label{fig:dalle2}
\end{figure}

The Flamingo model \cite{alayrac2022flamingo} is able to answer general purpose questions having as an input photos and text as we can see in Figure \ref{fig:flamingo}: 

\begin{figure}[h]
    \centering
    \includegraphics[width = 1.0\textwidth]{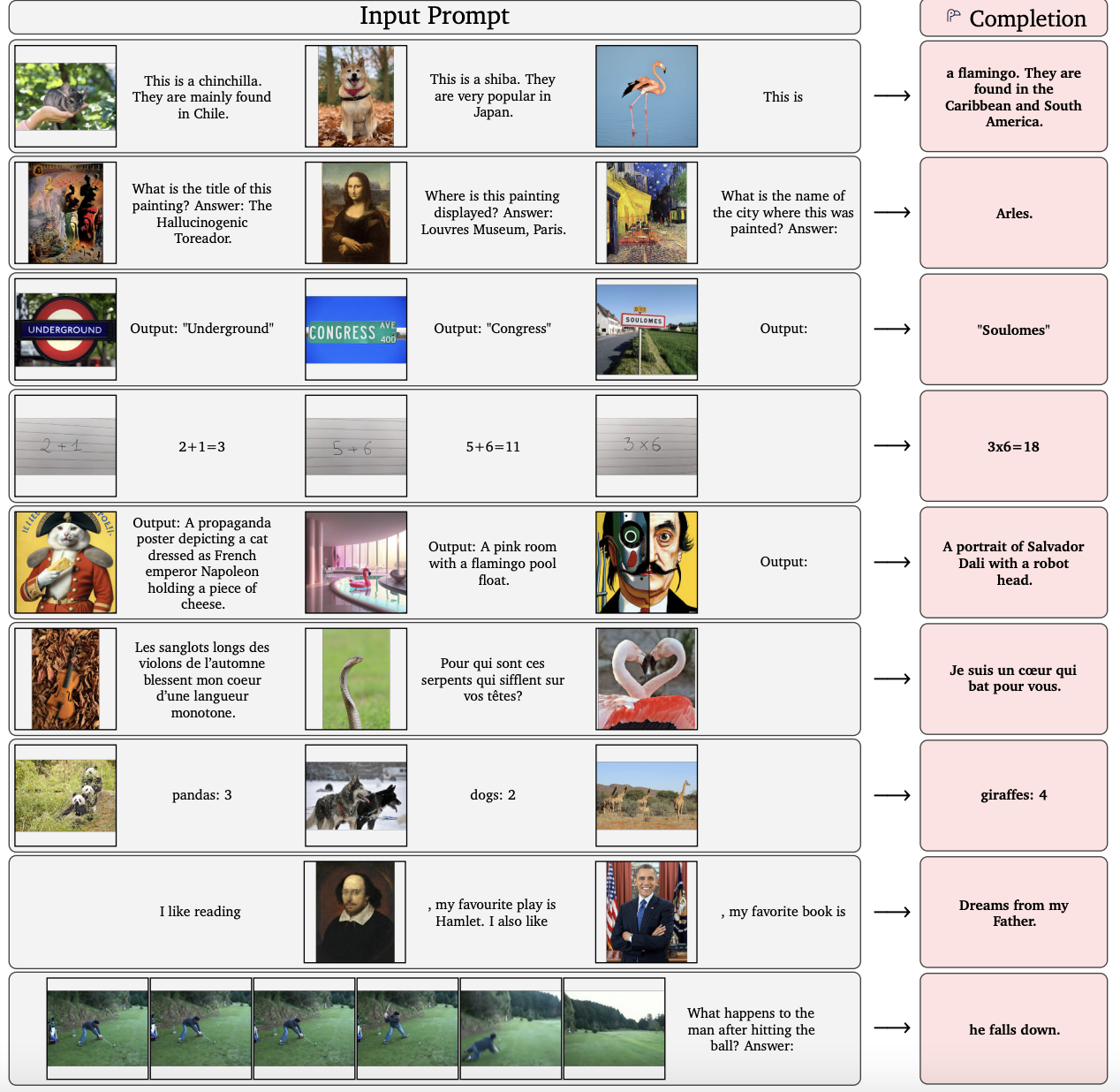}
\caption{Examples of input queries processed by the Flamingo model, at the left of the figure, and the outputs generated by the Flamingo model, at the right coloured in pink \cite{alayrac2022flamingo}.}
    \label{fig:flamingo}
\end{figure}

\begin{figure}[h]
    \centering
    \includegraphics[width = 0.8\textwidth]{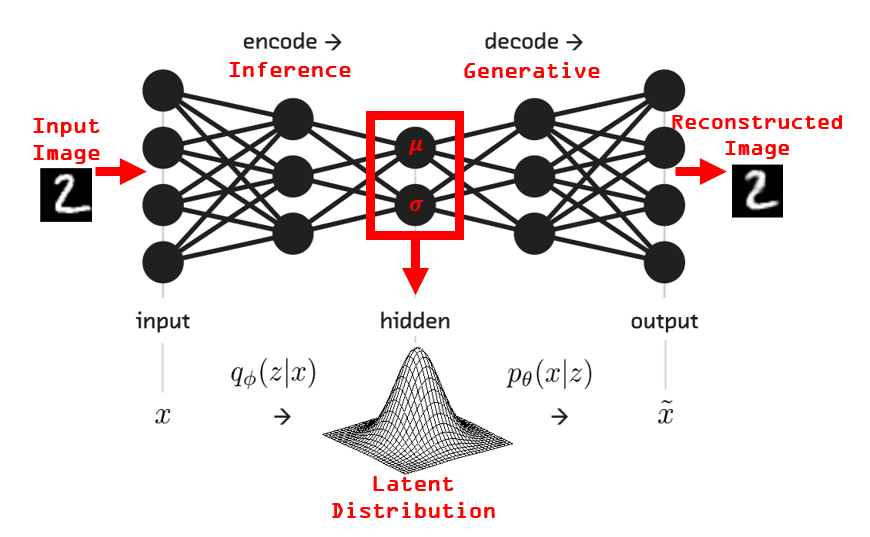}
\caption{Latent space of a transformer model that consist on Gaussian distributions that generate reconstructed images using sampling.}
    \label{fig:transformer}
\end{figure}

And lastly, another example is the GPT-3 model \cite{floridi2020gpt}, which can write texts in a particular style of a writer by processing its texts. There are more examples of such deep neural network models that are, generally, transformer models \cite{lin2021survey}. The intuition behind the transformer model, generally speaking, is that it has several encoder modules that learn a latent representation of the information such that several decoder models optimize the output, sometimes probabilistically, generated by the latent representation learned by the model. We can see a simple instance in Figure \ref{fig:transformer}. In that example, we only have a single encoder and decoder module and both the input and the output are photos. In particular, this is a variational autoencoder model \cite{zhai2018autoencoder}. 

In that example, the deep neural network learns the mean vector and covariance matrix of the multidimensional Gaussian distribution of all the number symbols that minimizes the errors of the reconstructed image. Most critically, the reconstructed image is sampled from the Gaussian distribution, so the numbers generated by the network are completely novel. Having studied this example, we can think about DALLE-2 or GPT-3 as more complex models that follow this logic using much more complex high-dimensional distributions to encode texts and generate, for example, images. 

\begin{figure}[h]
    \centering
    \includegraphics[width = 0.8\textwidth]{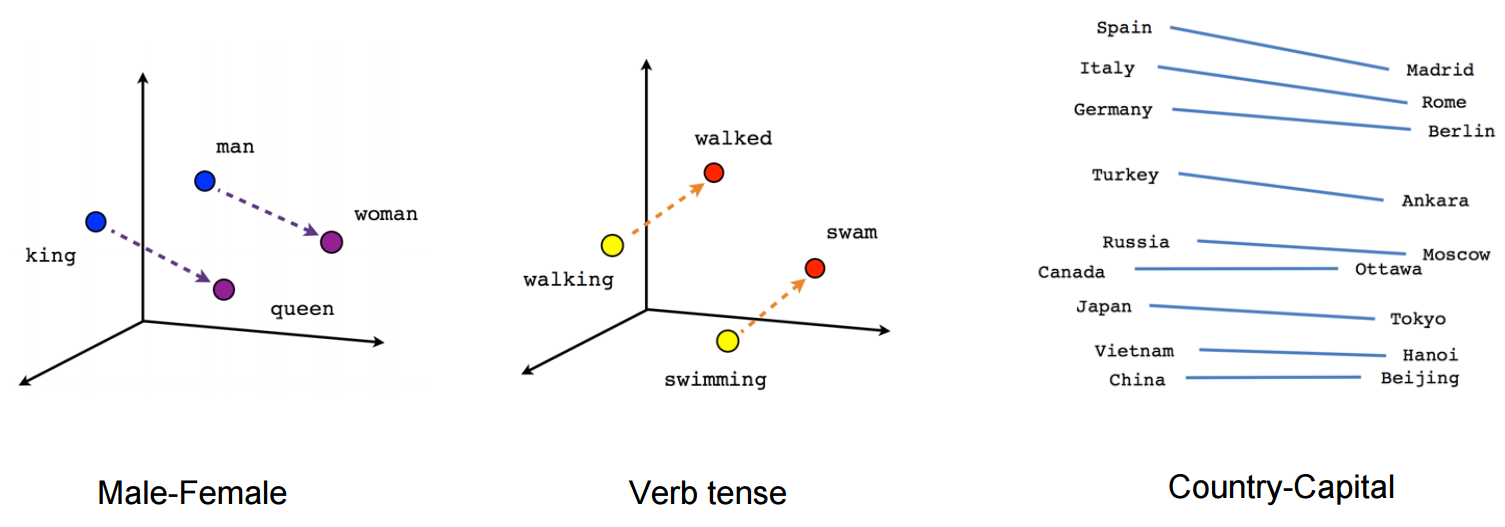}
\caption{Applying linear algebra to perform semantic operations over the latent space of a transformer model \cite{abadi2016tensorflow}.}
    \label{fig:tensorflow}
\end{figure}

Most interestingly, the latent space represent text or image information in a real high-dimensional space. Hence, we can use linear algebra to perform operation over the meaning of words, as described in Figure \ref{fig:tensorflow}. 

However, we argue that these models do not understand the meaning of the words. They are just learning a high, dimensional meaning space representation of the words. In particular, the meaning of those words is a quale (meaning a “quality,” a unitarily grasped element that is incorporated into an internal world of perception) that emerges from that representation and is perceived by human beings. This process of perception resembles what has traditionally been understood as “intuition,” or immediate grasping (i.e., internal apprehension) of an idea. Yet, up to which point could understanding as such be conceived without invoking a certain concept of subjectivity? If we are speaking in terms of internal assimilation, it is mandatory to elucidate the nature of the system that performs this task. Its “internal” world must incorporate the possibility of referring the object of perception to its own “inner” dimension, to its own “subjective core”. Indeed, this subjective information is not perceived by machines. Machines limit themselves to operate with signs, following a set of rules (be it clearly defined by instructions, be it inferred by statistical learning) that allows them to reach conclusions based on a set of premises and a set of rules of inference. As sophisticated as the process of instruction may be, and admitting to the possibility of designing a flexible set of instructions, in which machines may learn to learn by themselves, thereby reaching a higher degree of independence concerning the initial set of instructions, it is, in any case, a syntactic process, whose nature is sequential and algorithmic. In none of the cases so far described has the syntactic dimension been abandoned. A truly qualitative leap, leading to the semantic dimension, would demand articulating the possibility of grasping a meaning, displaying an “intuitive” behavior, thereby manifesting the existence of an internal world. Consequently, although the model represents the meaning of the words, there is no convincing proof that the program shows any distinctive sign of consciousness, without hard philosophical assumptions like multiple realizability and the ones summarized on the functionalism school of thought \cite{merchan2020machine}, “free intelligence”, and, hence, real understanding. Thus, even if complexity has increased in notable ways, the barrier between the syntactic and the semantic cannot be said to have been crossed (unless one admits some sort of mysterious “emergence” of properties, operating, virtually, ex nihilo, and without any plausible mechanism that may allow us to follow the sequence of steps leading to it, in order to explain how the process of emergence takes place). Obtaining higher degrees of complexity in the syntactic dimension may be a necessary condition, yet it is not a sufficient condition for reaching the semantic domain, in which real understanding may take place. We will return to this question in the following section.

\begin{figure}[h]
    \centering
    \includegraphics[width = 0.8\textwidth]{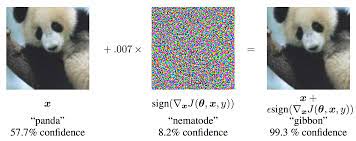}
\caption{Adversarial attack of a panda photo \cite{goodfellow2014explaining}.}
    \label{fig:panda}
\end{figure}

A demonstration of the claim mentioned in the previous paragraph (namely, that there is no need to invoke the concept of understanding, in its “strong” meaning, in order to explain the behavior of complex AI systems) is what adversarial attacks empirically show in neural networks. In particular, adversarial attacks try to add empirical evidence to verify the claim that these models do not understand the meaning representations that they learn. Concretely, the models are just minimizing a loss function that is estimated via the samples that the decoder modules generate using the information learned by the encoder modules. Figure \ref{fig:panda} shows an example of the adversarial attack, where a panda photo correctly classified by a model is perturbated by white noise, making the model fail its prediction of the panda photo and classifying it as a gibbon with $99.3\%$ confidence.  

\begin{figure}[h]
    \centering
    \includegraphics[width = 0.8\textwidth]{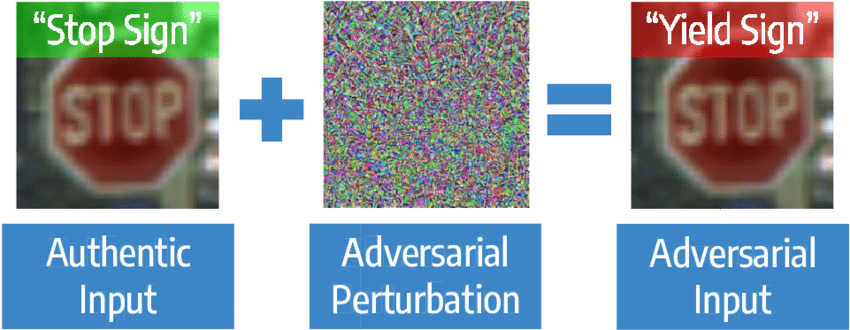}
\caption{Adversarial attack on a stop sign \cite{eykholt2018robust}.}
    \label{fig:stop}
\end{figure}

It is clear to human beings that the new photo is still a panda but the model fails dramatically. The cause of this behavior is that the model is learning spurious correlations about the panda but it does not understand the meaning of being a panda. Not being able to understand the meaning of an image may cause dangerous situations such as the one described in Figure \ref{fig:stop}.

Although adversarial attacks show that these models cannot understand by themselves, they also reveal that they could always grow in capacity to resist them, by being trained to avoid these attacks \cite{madry2017towards}. However, from a philosophical perspective, not being able to perceive the qualia of the representations implies a lack of understanding that makes the machine unable to appreciate, for example, the meaning of the images. From a statistical point of view, it is all about learning to fit curves, patterns, and correlation recognition in high dimensional probability distributions; a process that, once more, does not require any kind of allusion to the semantic dimension, and therefore to a strong conception of understanding, in which a being is endowed with mental states and can grasp the meaning of an object.

\subsection{Neurosymbolic artificial intelligence}

Machine learning and deep learning systems have been criticized as just performing complex curve fitting. Although some models can be considered universal approximators of functions, they do not understand the semantics of the representation nor the causal relations between the regarded variables of the problem.   

Motivated by this argument, the neurosymbolic community \cite{de2020statistical} has introduced algorithms that incorporate the semantics of the concepts represented in deep learning models through labels and symbols as in GOFAI systems, in an attempt at providing them the semantics that a deep learning model is unable to understand. In essence, neurosymbolic artificial intelligence is a hybrid of rule-based AI approaches with modern deep learning techniques.  

A popular, and classical, example of a neurosymbolic model is a sum product network \cite{poon2011sum}. In particular, sum-product networks are directed acyclic graphs whose variables are leaves, their sums and products are internal nodes and it has weighted edges. Although they share some structure with a neural network, even its parameters being optimized using gradient descent methods as in the case of deep neural networks, they encode additional semantic features such as the sums and products.   

\begin{figure}[h]
    \centering
    \includegraphics[width = 0.8\textwidth]{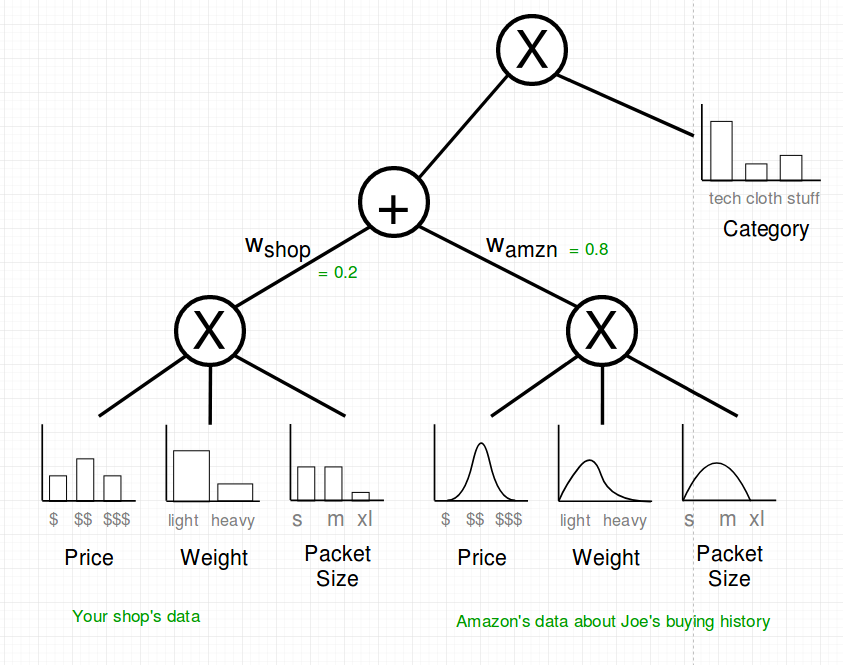}
\caption{Example of a sum product network dealing with prices of different providers such as Amazon. Sum product networks are able to introduce explicit meanings in its structure that can be interpreted as neurosymbolic artificial intelligence. However, even in models that introduce more rule-based AI combined with deep learning, semantics are introduced via syntaxis features.}
    \label{fig:sumproductnetwork}
\end{figure}

Nevertheless, from our point of view, those semantics are also a syntactical feature added to the deep learning model to make it execute a particular logic given a particular state of the neural network. Thus, it will have the same drawbacks as GOFAI systems, although it will, sometimes, increase the accuracy of the deep neural network for particular cases of dramatic failures. For example, it can be used to reduce racism bias or explicit violence. Another argument is that neither deep learning systems nor rule-based systems are aware of themselves, and therefore it is questionable to attribute a strong form of understanding (or “real understanding,” in its proper semantic dimension) to them. Hence, we can hardly expect any sort of magical emergence of consciousness from current neurosymbolic artificial intelligence systems.

\subsection{Causality models and probabilistic graphical models}

The previous models can represent knowledge of the real world and adapt to changes in context, as reinforcement or active learning algorithms do \cite{kaelbling1996reinforcement}, even if it is achieved by representing fixed semantics as syntax labels (like neurosymbolic AI does). However, causal relations are not targeted by any of these models. Causality models \cite{pearl2009causality} try to determine the causal relations that can be present in a probabilistic graphical model that represents the dependencies of random variables. Probabilistic graphical models, like Bayesian networks, are useful models to encode probability distributions over high-dimensional problems. They consist of multivariate distributions over large numbers of random variables that interact with each other.   

For instance, in the following figure we see an example of a probabilistic graphical model concerning lung cancer \cite{puente2017summarizing}. Each node represents a random variable of some event like a person that smokes. A probability distribution can be placed for every random variable and, if we sample that probability distribution, its result is introduced in another random variable that is causally linked by an edge in the graph. The whole probabilistic graphical model can be represented by a joint multivariate distribution. 

\begin{figure}[h]
    \centering
    \includegraphics[width = 0.8\textwidth]{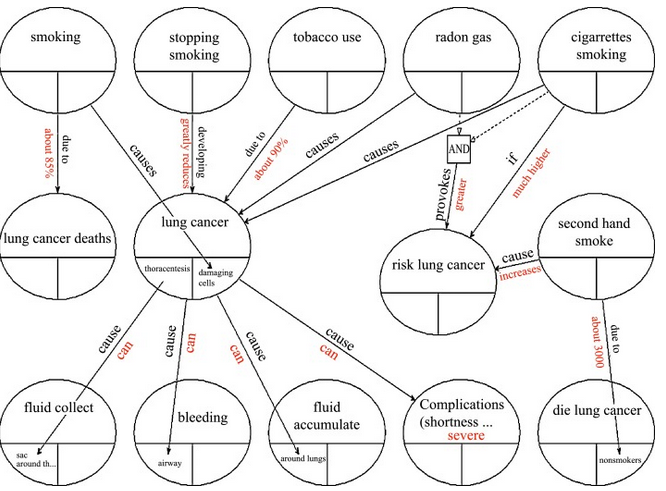}
\caption{Causal probabilistic graphical model (Puente, 2015) concerning the causes and effects of lung cancer. Each node represents a random variable and each edge represent the causal relation between the random variables. The whole probabilistic graphical model can be represented by a joint multivariate distribution.}
    \label{fig:causalpgm}
\end{figure}

Probabilistic graphical models have been used for causality introducing the do-calculus operator (Pearl, 2009). The intuitive idea is to deal with the fact that correlation is not causation in this framework. The do-calculus operator consists of conditioning a random variable to a particular observed value of it. For example, if we perform a real experiment and we observe that smoking causes lung cancer with a certain probability, we can condition the dichotomous random variable smoking to true. Hence, smoking is not anymore a random variable but a value, and the causal relation is represented by the graph.   

Although causality has been widely applied in probabilistic graphical models, the model still does not understand what it knows or does not know. In essence, probabilistic graphical models encode causal relations and a multivariate random variable; yet, they are representation models, not conscious models. Hence, they are not able to understand its representation, regardless of the complexity of their design. 

It is important to remark that the previous models represent reality as given knowledge and as a predefined architecture. However, they do not integrate a semantic perception of its representations. Indeed, one requires an external observer that perceives the qualia of the representations.   

\section{Philosophical discussion}

The second section of this paper deals with the nature of understanding and its essential role in addressing the question of the existence of intelligence in the so-called intelligent machines.   

In this section, we will develop the idea of the non-computability of understanding (following, among others, \cite{penrose1994shadows}). In our view, a satisfactory concept of understanding, induced from the phenomenology of human understanding (and possibly from other forms of animal behavior), requires the grasping (as internal assimilation, or meta-representation) of an object, beyond the mere manipulation of signs. Indeed, it requires ascribing meaning to a sign and therefore having an intuition of a semantic content that cannot be reduced to the syntactic processing of instructions in an algorithmic way (and therefore in a finite number of steps), as a point of departure fixed from outside, by an external agent. Yet, this process cannot be explained without presupposing some kind of subjective or internal “instance” in charge of apprehending such a meaning.   

Instructions (which, for the sake of simplicity, sometimes we shall include under the general philosophical labeling of “rules”; the concept points to the presence of an initial determination, as flexible as it may be –indeed, it can be the result of statistical inference rather than direct programming-, opposed to the possibility of self-determining its own actions) can be defined as part of a syntactic domain, in which a set of orders allows the system to make transitions from premises to conclusions, following relations of logical consequence (If p then q is the basic rule of inference in a set of well-formed formulae). These instructions may offer direct rules, or they may generate the possibility that the system extracts patterns through models whose parameters are inferred by the processing of data or statistical inference (rather than simply relying on a strict, clearly defined, set of rules, as in old artificial intelligence). Yet, even if these systems manage to learn to learn, rather than following directly fixed rules, and they are capable of generating their own instructions “from inside”, their observed behavior still adheres to an algorithmic processing of information. Thus, they can be interpreted as examples of a universal Turing machine, as we shall discuss.

Understanding, in its “strong” sense, cannot consist merely in the syntactic dimension, in the manipulation of sequences of signs in a finite number of steps (which, so to speak, would allow us to adopt a “mechanistic approach” to the entire process, where understanding would be reduced to a sequential arrangement of elements). Rather, from a phenomenological point of view (and therefore from its observed manifestation in a human mind –and possibly in other forms of animal mind--) understanding  implies ascribing meaning to a sign, and therefore the possibility of having an internal representation of that representation (the sign itself), in which the subject becomes aware of his representation. It is doubtful that AI, in its present stage of development, has reached the level of understanding, which would belong to a semantic domain, in which internal awareness permits to attribute meaning to symbols.  

The problem of subjectivity is intimately connected with this question. Indeed, understanding, as a “non-blind processing of signs”, demands the possibility of having internal representations, i.e. mental states in which an agent is capable of referring representations to himself, to his internal structure, so to speak; id est, to a subject, or “pour soi.” How can this be comprehended without attributing a faculty of “intuition” to that agent, and therefore a nature similar to that of a subject, as is phenomenologically observed in the case of human beings? In a purely behavioristic approach, this notion of understanding will seem superflous. Indeed, this is the case we want to make: that understanding, as internal grasping of a meaning, is not necessary for explaining the behavior of AI machines. Whether it is necessary for explaining human behavior is a problem beyond the scope of this paper; we suspect that the answer is in the positive, yet, regardless of the answer given to it, the idea of understanding that we discuss here is more demanding, philosophically speaking, than a merely sequential, mechanistic approach, and it clearly points to a stronger conception of understanding as such, which would incorporate a potential subjective stance.  

The duality between rules (once more, understood in a generic philosophical manner, encompassing direct rules through instructions and a more flexible design, in which the system infers its own rules from statistical pattern recognition –thus, “learning to learn” by itself--) and intuitions, or between syntaxis and semantics, seems irreducible in the present stage of human knowledge. We cannot explain a hard form of understanding, of “subjective” apprehension (of “realization” of what is going on there, by being aware of meanings instead of simply using rules), without invoking the existence of the power to intuit. It may even respond to a fatality of human reasoning (i.e., to a “transcendental impossibility”, in the Kantian sense, derived from the structure of our understanding). Yet, it does not imply that AI will never achieve the goal of creating a hard form of intelligence, one in which the machine is not only capable of processing information but also of apprehending meanings and therefore of understanding the “hard interpretation” of the term.   

So to speak, if the processing of information can be regarded as “first-degree” assimilation –-given that the machine needs to learn a set of rules or instructions--, the subjective assimilation of that information in terms of understanding, or intuitive grasping of meaning, is a second-degree assimilation, or an internal “formalization,” according not to a blind set of rules, formally defined –-thus, syntactically--, but to its own subjective rules (to its own “intuition”). As sophisticated as the design of the system may be, and even if the system is not set to blindly obey an initially set number of instructions (as is the case with GOFAI, in contrast to neural networks, in which there is statistical pattern recognition and the system is capable of, so to speak, “self-designing” itself), we are still under the domain of syntaxis, without real awareness of what’s going on there, and thus without the possibility of attributing meanings to symbols. 

An important question surrounding this problem can be framed in the following way: how is information represented?  

If we constrain our analysis to information represented by the human mind, as is well known there are at least three main schools of thought. The initial paradigm in the cognitive sciences understood the human brain as a processing information machine, ruled by an internal symbolic code, or internal "language" (what Fodor has named "mentalese"; \cite{fodor1995elm}). A single, unified code of representation would therefore underlie the brain's ability to represent information about the external environment and its internal milieu. The limits of this paradigm, however, in particular its lack of specific neuronal translation, led to the development of a second paradigm. According to it, generally called "connectionism", there is no single code in the brain. The different input/outputs of a vast network of neurons processing in parallel are the way in which the brain stores and manages information. Of course, the creation of artificial neural networks is based upon this paradigm. Thus, its power to establish fruitful bridges between neurosciences and artificial intelligence has been one of the keys to its success, taking into account that it contributes to reinforcing the computational view of the mind.   

In any case, the weaknesses of this paradigm are also worth noticing. In particular, does it really explain understanding as such? We are still within the domains of "information processing," yet it is not clear that this approach to the nature of mental activity, as powerful as it may seem, incorporates a convincing theory of how a meaning is subjectively apprehended --i.e., how we understand anything at all.  

An additional paradigm is that of "spatial models" \cite{gardenfors2014geometry}. According to it, cognition can be modelled in topological and geometrical spaces. Inspired by geometrical categories, the paradigm suggests that information is organized in spatial structures representing "meanings" through connections (like relations of proximity, convexity, coordination...). Thus, one may speak in terms of a real "geometry of meanings". In a more advanced development, this approach has been linked to some promising neuroscientific results concerning the encoding of spatial information by the brain. As Bellmund et alii have written, "place and grid cells can encode positions along dimensions of experience beyond Euclidean space for navigation, suggesting a more general role of hippocampal-entorhinal processing mechanisms in cognition" (Science 362 2018). 

Yet, the question as to the subjective formation of meaning remains essentially unanswered. All these approaches may attack specific dimensions of the problem. However, the do not solve the hard problem of meaning: what is it to understand? How do neural processes generate understanding? This question is intimately connected with the problem of attributing understanding to a computer, and therefore to the question concerning the nature of mental states.  

The construction of a prototype through category spaces does not elucidate how we grasp meaning. Information processing through a set of instructions (given by design or spontaneously generated by statistical inference) does not exhaust the conceptual problem of comprehending a meaning, because subjectivity is inevitably involved in the analysis. One can, indeed, develop sophisticated models in terms of conceptual spaces and the assignation of elements to "prototypes", thereby establishing connections between different objects; yet, how do we "conceive"? How do we assimilate an internal representation, even if, externally, it can be visualized as a set of prototypes situated in conceptual spaces? 

The grasping of the meaning of a concept is a remarkable activity. As is well known by experience, children are capable of understanding the meaning of certain words only after being exposed to a few examples. Neural networks, on the contrary, need a much higher number of examples to acquire a stable representation. How does the human mind manage to generalize, and how does it so quickly? How do we get to understand words in such rapid manner? Moreover, how do we represent and understand logical constants ("and", "or", "if"...) and pure abstractions (totality, nothingness...)? 

Even if one can connect, on solid grounds, conceptual spaces with meaning, and words topologically mapped to a potential interpretation, the precise way in which this is done remains a mystery. The gap is too deep. We still do not know not only how psychological spaces of representation are related to the neural organization of the brain, or, why not, to the circuitry of a computer, but also how such psychological spaces are referred to our subjective experience, to our "understanding".  The problem of finding the exact correspondence between mind and brain is still present, because the creation of psychological spaces cannot be examined only "objectively", externally, by representing it in terms of, for example, conceptual spaces: it also needs to address the internal dimension of the process, the way in which the subject becomes conscious of that representation, so that it does not merely consist of an arrangement of symbols, but of a meaningful experience. Mental maps are a useful tool to organize our understanding of how humans understand, but they do not explain the process of understanding itself, of "intuiting" a meaning, of "eureka".  

The previous considerations can be summarized in the following tenets, which, in practical terms, work as axioms, induced from the phenomenology of how human understanding appears to operate: 

\textit{"If a machine is capable of understanding (in the strong sense), then it must be capable of combining rules and intuitions"} 

\textit{"“If semantics cannot be reduced to syntaxis, then a machine cannot understand"}

It is clear that, if we accept these two tenets (which, once more, can be seen as definitions, or as ex hypothesi projections; however, one must bear in mind that they are not arbitrary, as they respond to legitimate inductions from phenomenological observations), in the present state of AI there is no legitimacy in stating that the so-called intelligent machines understand anything (and therefore that they are truly intelligent, in a psychological sense, which transcends mere task-solving, as it demands the existence of an internal mental state); they simply process information. They rest in the syntactic domain, without penetrating into the semantic domain. This thesis is, certainly, coincident with Searle’s famous Chinese room experiment \cite{searle1982chinese}; in our view, the system behind the door offers no sign of understanding Chinese, or of understanding anything at all. The process is not significant for itself, because there is no internal world to which this significance could be referred. Such internal world is equivalent to the traditional philosophical concept of “für-sich”, of “for-itself”: of a point of view, of a “subjective experience” to which the external –the element of perception—can be referred. As Thomas Nagel has written, “let me first try to state the issue somewhat more fully than by referring to the relation between the subjective and the objective, or between the pour-soi and the en-soi. This is far from easy. Facts about what it is like to be an X are very peculiar, so peculiar that some may be inclined to doubt their reality, or the significance of claims about them. To illustrate the connection between subjectivity and a point of view, and to make evident the importance of subjective features, it will help to explore the matter in relation to an example that brings out clearly the divergence between the two types of conception, subjective and objective. It is not equivalent to that about which we are incorrigible, both because we are not incorrigible about experience and because experience is present in animals lacking language and thought, who have no beliefs at all about their experiences” \cite{nagel1974like}. 

Nevertheless, it is necessary to deepen into the question concerning the nature of understanding. Some sceptics may argue that we have deployed such a demanding notion of understanding that, ex hypothesi, a machine will never be able to understand, given that it will never be able to display subjective thinking. By setting so high a barrier, this philosophical approach to understanding would prevent any computational architecture from ever achieving strong understanding as such.  

In general terms, thinking can be defined as an association of mental contents. This approach to thinking can be applied not only to humans, but to any biological species capable of forming internal representations of the world, and therefore of having mental states. To think coincides with the act of analyzing and selecting mental contents; with an internal filtering of options in which a multiplicity of possible combinations gets reduced to a specific “piece of thought.” This relation of ideas must be expressed in a language. A language appears as a system of signs, useful for a certain agent for whom this set of signs is significant. The feature of language as a system of signs has been encapsulated in Chomsky’s famous definition of language as “a set of (finite or infinite) sentences, each finite length, constructed out of a limited set of elements” \cite{chomsky1957syntactic}. 

Chomsky has insisted on the difference between rules and representations in language \cite{chomsky1980rules}, a distinction that can be extrapolated to consciousness (and understanding as the act of a conscious subject). A competent agent must be capable of representing meanings, rather than simply following direct rules or patterns statistical extracted from data. It is questionable that there would be real communication (internal or external; with oneself or with others) without the possibility of understanding meanings, and therefore of intuiting a certain mental content, represented in that language.  

Understanding requires the assimilation of syntactic sequences; the perception of a logic referred to the subject, to its internal dimensions. The association of mental contents that underlies the phenomenon of understanding would be unintelligible without, precisely, the presence of a “mind”, of a mental state capable of attributing meaning to a syntactic sequence. The thinking subject assimilates a logical sequence by, so to speak, designing a “function of categorization”, which interprets the different elements at play in such mental association. Expressed in modern terms, this is essentially equivalent to the Kantian perspective. According to the philosopher of Königsberg, all acts of the understanding can be reduced to judgements, so that the understanding can be represented as a faculty of judging \cite{kant1908critique} (B94).The Kantian categories are a faithful expression of the way in which the understanding (human subjectivity) processes the phenomenal multiplicity. “The I-think (ich denke) expresses the act of determining my existence” \cite{kant1908critique} (B157): the act of thinking determines the subjectivity of the knower. Every act of understanding or sensitivity must be accompanied, as Kant says \cite{kant1908critique} (B132), by the I think; an intuition that is a representation prior to all thought. Once more, we return to the problem of subjectivity and, consequently, to the mind-body problem as the overall conceptual difficulty haunting the whole question about the possibility of constructing thinking machines. Who would be thinking, after all? Who would be understanding? Would the machine be the subject of anything? Would it possess an internal dimension, capable of controlling, as a supervising instance, the processing of information?  

Critics may argue that this replicates the famous “ghost inside the machine” fallacy (which incurs in a homunculus problem, and therefore in a self-reflexivity paradox, expanded ad infinitum), but even if one adopts a clearly behavioristic account of mental processing, it is hard to understand how to explain the phenomenon of comprehension, of the assimilation of a meaning. This “meta” framework, by which the mind supervises what if has before itself, is conceptually inseparable from the assumption of an internal dimension, of a “Self”. 

 Generically, the act of thinking appears as the connection between mental contents through logical and grammatical constants. This phenomenon can also be visualized as the design of a function with an application domain: that of the objects on which that thought deals. However, this analysis of the fundamental characteristics of thought would be incomplete without distinguishing between the perspective of rules and that of intuitions \cite{blanco2020integration}. In its most algorithmic or regulated facet, rational thought is structured by rules that guarantee the possibility of reaching consistent conclusions, and the thinking subject must show competence in the use of those rules. However, this thought must be supervised by a subject that assimilates the contents and is capable of apprehending a meaning, a unitary sense of the set of mental contents that constitute that specific thought. This assimilation of the object as such (be it a concept, a principle of reason or the integration of both within a proposition), evaluated in its unitary dimension and not only in that of the individual elements that constitute the object, seems to evoke a genuinely "intuitive" facet of the mind, where the analytical decomposition of the parts that come into play in the content of thought gives way to the elaboration of a unitary synthesis. Therefore, following syntactic rules or extracting them from statistical inference is not enough: it is necessary to take charge of them, even if precariously. These considerations do not imply, however, accepting a kind of unilateral and despotic primacy of the intuitive over the rational, as if elements inaccessible to a logical and scientific understanding appeared in thought. It is not magic. In fact, the contents of our intuitions have to be subsumed, in one way or another, in the general mechanisms of rational thought, in the rules that guide our intellectual processes. It would not be an exaggeration to maintain that intuitions obey some sort of “internal rules,” so the dream of explaining how human thought works would not be entirely banned from the scientific efforts of humanity.   

 \section{Conclusions}
In the first section we have exposed and evaluated the present models of AI. Our critical assessment has included the problem of conscious representation, of the existence of mental states, and of the presence of real intuitions beyond the processing of information. We have tried to show that the AI systems, at their present stage of development, are incapable of grasping anything at all, as they cannot refer a certain representation to their own “self”, or internal dimensions (to their “subjective core”). This dimension, which have labelled, rather generically, as “semantic” in opposition to “syntactic” (which would encompass the processing of information, be it by following a set of rules or by inferring them from statistical patterns), is, in essence, non-computable, intuitive, and subjective.

According to the evidence on the current state of AI, a syntactic scheme suffices to explain the behavior of the machine. Its observed behavior can be reduced to the processing information, after following a set of rules or after inferring it from its own process of statistical recognition of patterns. Thus, it would not be necessary to invoke a semantic dimension, of true (“strong”) understanding. By economy of hypothesis, it is then simpler to explain the observed phenomena without presupposing understanding in the machine, and therefore without presupposing some form, however elementary, of subjectivity (and, therefore, of the possibility of displaying autonomous behavior in decision making, stemming from the existence of an underlying subjective instance, of an I in charge of deliberating and acting). This is not an obstacle to the possibility of some kind of emergence of subjective properties in highly complex systems, capable of “learning to learn” by inferring their own instructions. Nevertheless, this thesis is something that, at the present time, cannot be assessed with certainty. Yet, in the present state of development, AI models offer no clear sign of subjectivity. It is not necessary to postulate the existence of a mind inside the machine, and thus it is not necessary to assume mental states, and therefore subjectivity, in the so-called “intelligent machines.” 

 In our view, the exhibited behavior of machines can still be explained by a mechanism similar to that of cascades of Estimuli-Responses. We are facing a strictly computable process, since we can follow the precise itinerary between the instruction and the response displayed, whose stages are finite in number. The fact that these instructions are so flexible as to allow the machine to reach its own conclusions after a processing of statistical inference does not invalidate our point. Moreover, in the case of human understanding it seems inevitable to introduce the self-referential dimension, the idea of an internal world, of mental states in which it makes sense to speak of sensation and understanding, and not of mere information processing, because there is a subject to whom we refer the process that we analyze as an objective phenomenon. 

One could reply that our approach is biassed by apriosim. By automatically defining understanding as we do, it is inconceivable that a machine could ever achieve it. Certainly, our idea of understanding is strong and intuitive, but it is also empirical, induced from the phenomenology of the process. It is born from the observation of how we humans (and probably other animals) understand. It cannot be discarded that we may discover a mechanism to imitate subjectivity, and build mental states that make this self-referentiality possible. Only future research will be able to clarify this transcendental question. What seems out of doubt is that scepticism about the existence of strong AI is justified. The question, however, points to the possibility of designing systems in which the “computable” may be merged with the “non-computable,” and the processing of information may be integrated with some form of subjective assimilation.  

\bibliographystyle{acm}
\bibliography{notes}

\end{document}